\documentclass{article}
\usepackage{spconf,amsmath,graphicx}
\usepackage[utf8]{inputenc} 
\usepackage[T1]{fontenc}    
\usepackage{hyperref}       
\usepackage{url}            
\usepackage{booktabs}       
\usepackage{amsfonts}       
\usepackage{nicefrac}       
\usepackage{microtype}      
\usepackage{xcolor}         
\usepackage{graphicx}
\usepackage[numbers]{natbib}
\usepackage[shortlabels]{enumitem}

\title{High-Fidelity Human Avatars from Laptop Webcams using Edge Compute*}
\name{Akash Haridas \quad Imran N. Junejo\thanks{*Patent pending.}}
\address{Advanced Micro Devices (AMD), Canada. \\
\texttt{akash.haridas@amd.com, imran.junejo@amd.com}}
\begin{document}
%
\maketitle
\begin{abstract}
Photo-realistic human avatars have broad applications, yet high-fidelity avatar generation has traditionally required expensive professional camera rigs and extensive artistic labor. Recent research has enabled constructing them automatically from smartphones with RGB and IR sensors, however, these new methods still rely on high-resolution cameras on modern smartphones and often require offloading the processing to powerful servers with GPUs. Modern applications such as video conferencing call for the ability to generate these avatars from consumer-grade laptop webcams using limited compute available on-device. In this work, we develop a novel method based on 3D morphable models, landmark detection, photorealistic texture GANs, and differentiable rendering to tackle the problem of low webcam image quality and edge computation. We build an automatic system to generate high-fidelity animatable avatars under these limitations, leveraging the compute capabilities of AMD mobile processors.
\end{abstract}
\begin{keywords}
AR/VR, Edge Compute, Face Avatar.
\end{keywords}

\section{Introduction}
\label{sec:intro}

The ability to generate photo-realistic human avatars from a small set of images has vast applications. For example, in video conferencing applications, a user can have the option to transmit their animated look-alike avatar mimicking their facial actions instead of transmitting their actual video feed, which can help preserve their privacy. Transmitting the avatar and its animated expressions instead of live video also results in significant bandwidth savings. Personalized avatars can also be used in video games and immersive virtual reality environments. 

For these use cases to be successful, it is important to generate avatars from a small set of user images captured in natural environments, such as at their work desk, using commodity cameras, like laptop webcams, using a limited compute available on-device. Traditionally, the production of virtual avatars has required advanced camera rigs, controlled lighting environments and artistic labor, and hence has been primarily limited to professional studios. Recent research in 3D reconstruction has enabled generating these avatars automatically from smartphones using RGB and IR sensors. However, these new methods still rely on high-resolution cameras and depth sensors on modern smartphones and often require offloading processing to powerful servers with GPUs. 

\begin{figure}[t]
	\centering
	\includegraphics[width=0.5\textwidth]{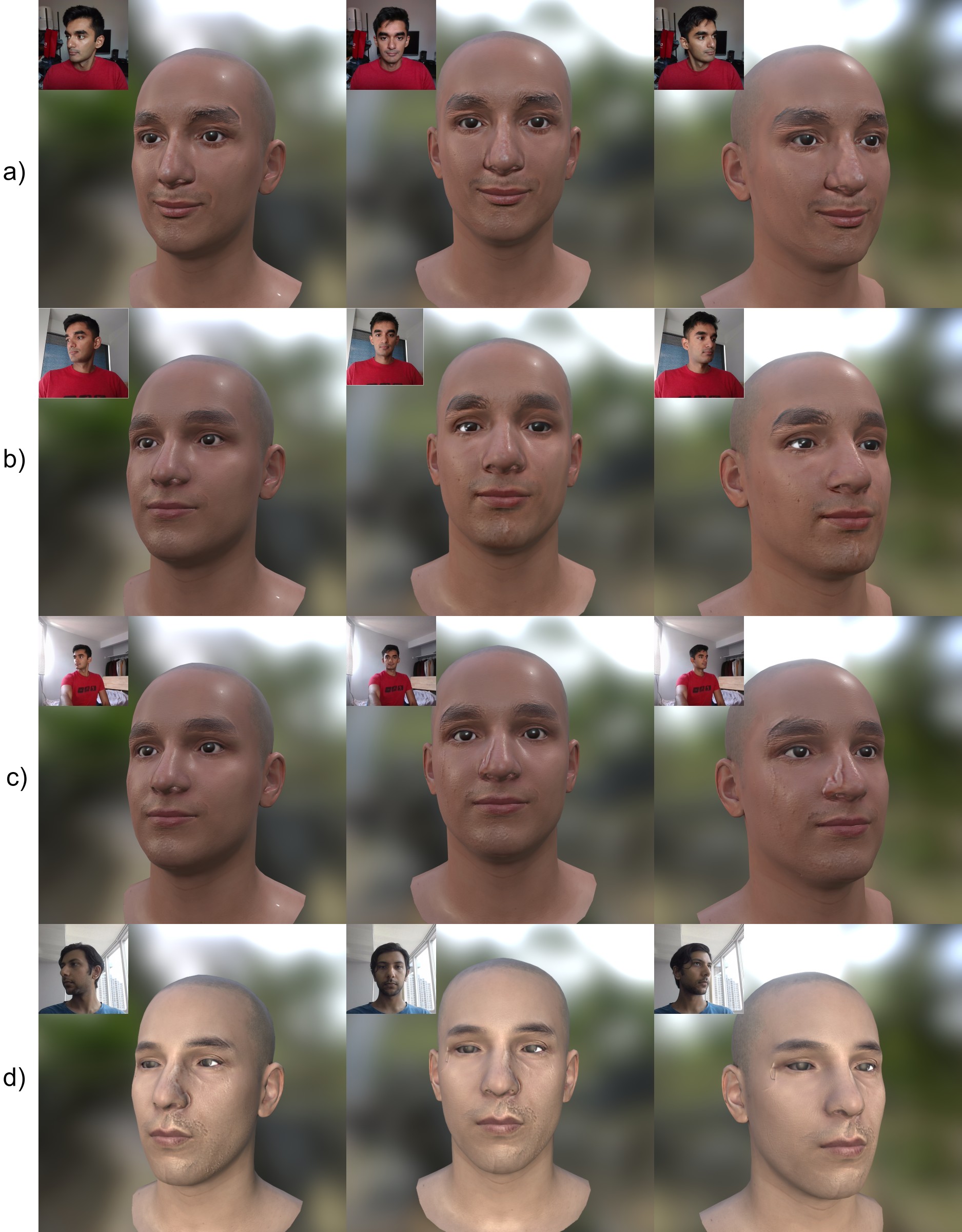}	
	\caption{High-fidelity renderings of the avatars generated from the multi-view image test sets captured on a laptop webcam.}	
	\label{fig:results_main}
\end{figure}

\begin{figure*}[ht!]
	\centerline{\includegraphics[width=0.98\textwidth]{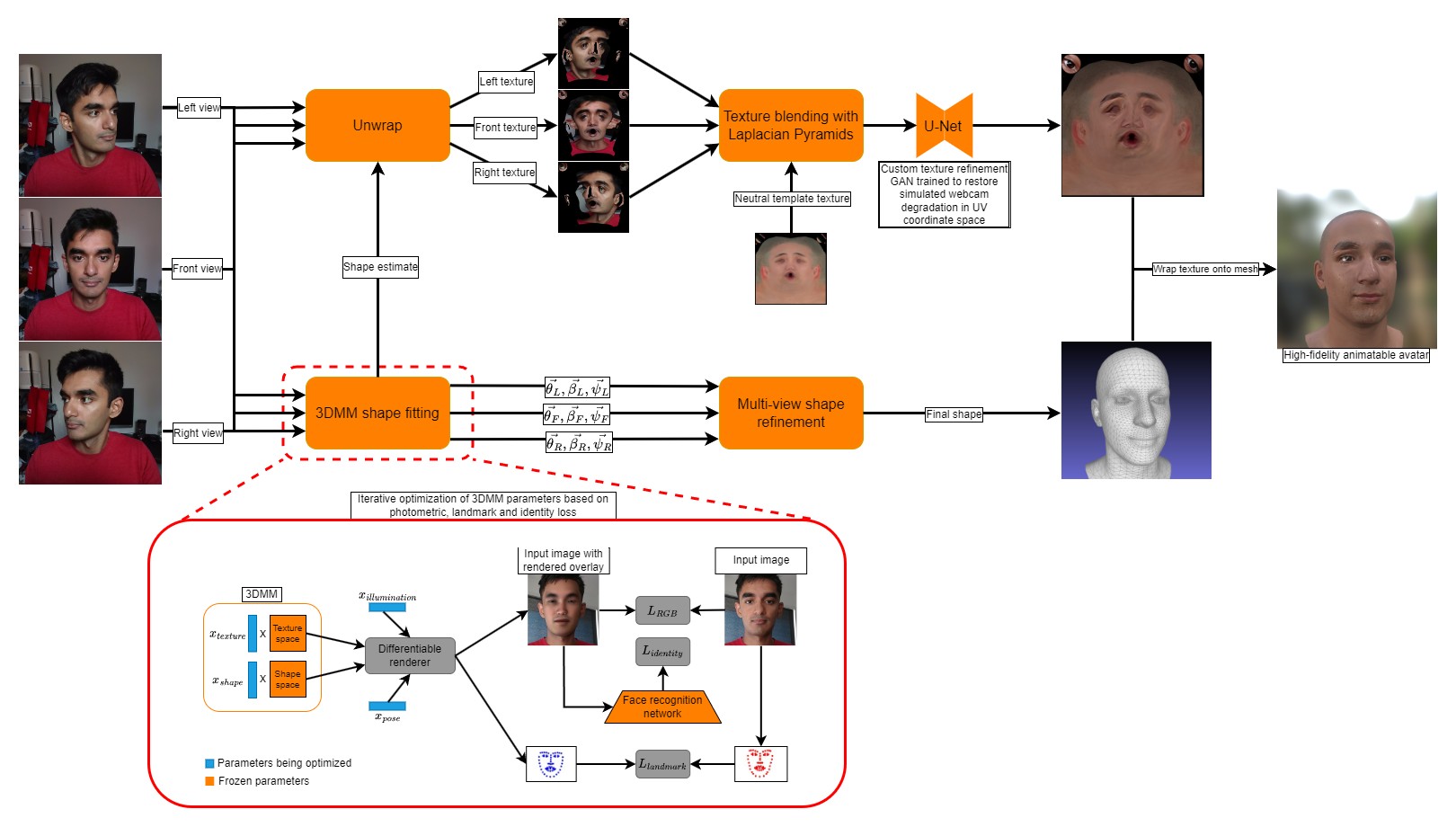}}
	\caption{Our end-to-end avatar generation pipeline}	
	\label{fig:pipeline}
\end{figure*}

In this work, we build an automatic pipeline to generate high-fidelity human avatars, with the following constraints:
\begin{enumerate}
	\item The pipeline must work with consumer-grade low resolution and/or noisy (1080p) laptop webcams that capture limited skin details and texture.
	\item The pipeline must handle images captured in natural, unconstrained settings without needing the user to modify their environment to be more suitable for photo capture. The pipeline should take images that may contain uneven, directional illumination across the user's face and output an avatar that is neutrally illuminated for use in downstream applications.
	\item The pipeline must run on-device on laptop hardware and generate the avatar in a few minutes.
	\item The resulting avatar asset must be readily animatable in downstream applications, rather than being a static 3D model.
	\item In order to capture a user face properly, we assume the pictures capture a user's face from multiple angles. \end{enumerate}

\noindent \textbf{Background}:
Prior work on automatic 3D reconstruction of human faces from a single or a few images is based on 3D morphable models (3DMMs) and implicit neural representations (such as NeRFs), or a combination of both. 

A \textit{3D morphable model (3DMM)} is a vector space representation of a class of objects (in this case human faces) - a set of basis shapes and basis textures, built from detailed 3D scans of a large and diverse set of human faces \cite{3dfmmpastandfuture}. Linear combinations of the basis vectors can synthesize new faces, and all linear combinations within a certain range are guaranteed to produce realistic faces. 3D reconstruction using 3DMMs usually involves estimating the parameters (coefficients) of the 3DMM that best explain a given set of images of a person's face by minimizing energy terms based on facial landmarks, pixel colors, or depth information using gradient-based optimization \cite{hifi3dface, deep3drecon, ffhq-uv}.

Considering the constraints detailed in the previous section, we choose to adopt the 3DMM approach over NeRFs based methods \cite{occupancy_networks, deepsdf, nerf} for ease of animation and editing \cite{blendshapes,nerfblendshape,nerface, nerfblendshape, avatarmav,instantngp,lugaresi2019mediapipe}, and the computational cost for both generation and rendering.


\section{Method}
The method consists of two parts: first, shape generation by fitting 3DMM shape parameters, and texture map generation (cf. Figure \ref{fig:pipeline}). Each sub-component has been designed to tackle a specific requirement of our target use-case and the pipeline as a whole is designed to be computationally efficient.

\begin{figure*}[!t]
	\centerline{\includegraphics[width=0.95\textwidth]{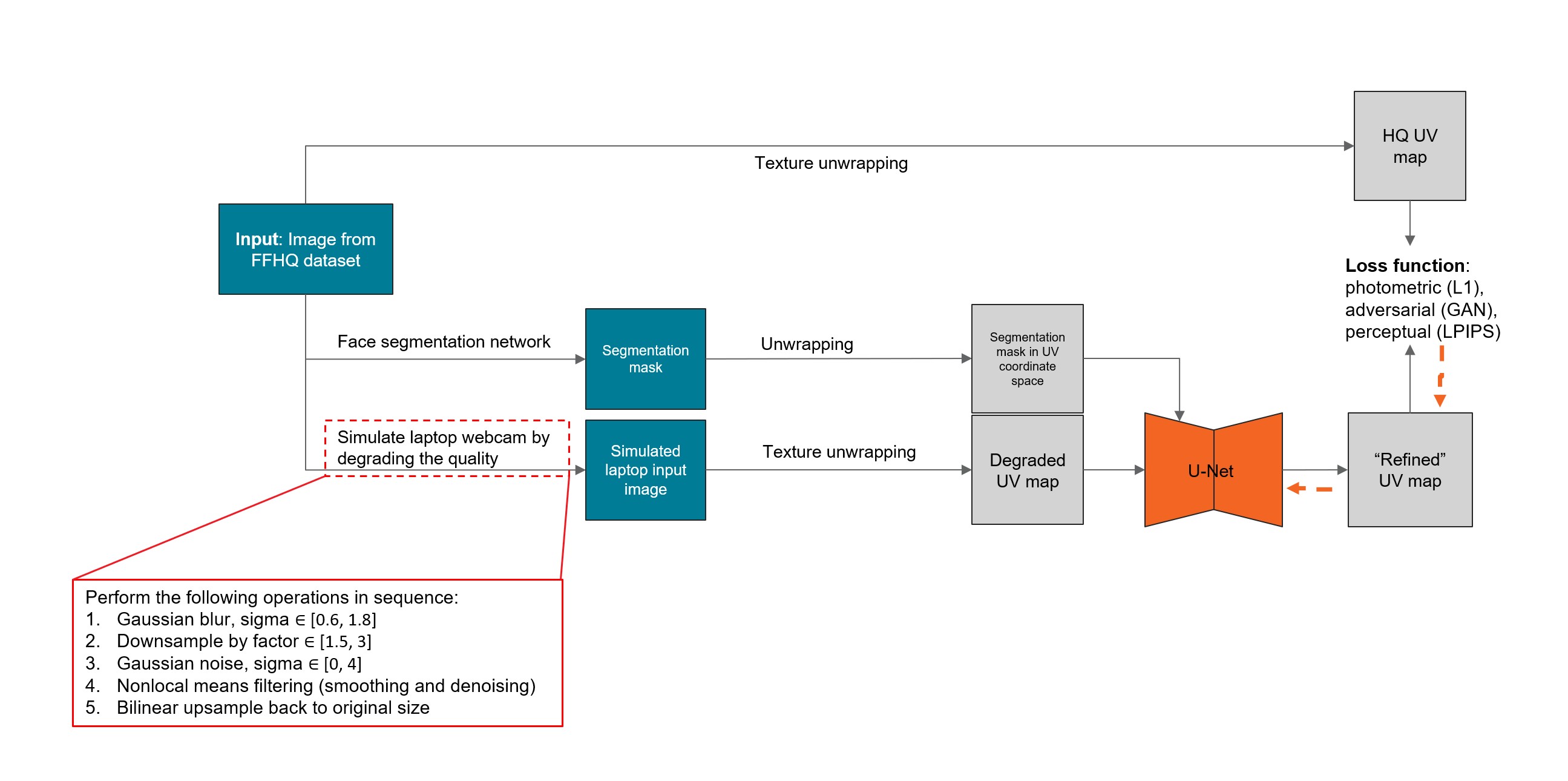}}
	\caption{The self-supervised pipeline to train a U-Net model to restore a simulated laptop webcam degradation in UV coordinate space. The orange arrows depict the backward propagation of gradients.}	
	\label{fig:texture_refinement_network}
\end{figure*}

\noindent \textbf{3DMM shape fitting:} Through iterative optimization, we fit the parameters of FLAME \cite{flame}, a 3DMM built on a large and diverse collection of head scans. As shown in Figure \ref{fig:pipeline}, the procedure starts with initializing the 3DMM shape and texture parameters as well as auxiliary parameters for head pose and scene illumination used in rendering the head. The shape parameters further consist of identity and expression parameters. The current estimate of the avatar is "decoded" using the 3DMM shape and texture spaces and rendered through a differentiable renderer. Several loss terms are calculated between the rendered image and the source image. $L_{RGB}$ computes the L1 distance between the RGB pixel values. $L_{identity}$ computes the L2 distance between the identity embedding vectors extracted from a pre-trained ArcFace face recognition model. $L_{landmark}$ is computed by detecting 105 facial landmarks on the source image. Together, this objective function enforces shape similarity via landmarks, identity similarity via embeddings from a face recognition model, and color similarity via the pixel-wise comparison. The 3DMM and auxiliary parameters are updated using the LBFGS optimizer. The optimization loop exits when the parameters reach convergence which takes around 200 iterations in our experiments.

When optimizing over multi-view images of the user, we initialize separate parameters for each image, except identity parameters where we initialize 1 set shared between all of them. Note that we also estimate a texture using the 3DMM texture space, since this is required for the rendering used in the RGB and identity losses. We discard this texture at the end of the optimization and instead generate a more high-fidelity texture using the procedure described in subsequent sections.

\noindent \textbf{Multi-view texture blending:} 
Having $N$ 3D meshes (with a shared identity), using aligned landmark vertices, we produce $N$ UV texture maps by projecting points on the 3D surface onto the aligned 2D image. Each of these $N$ texture maps cover a different portion of the user's head. Each of the source images will have some portion of the head that is not visible due to self-occlusion. The final avatar should have a seamless texture covering all sides of the head which we can obtain by blending together the $N$ unwrapped textures. The blending procedure, based on the methods of \cite{hifi3dface} and \cite{ffhq-uv}, is guided by binary facial segmentation maps estimated from the source images using a pre-trained face segmentation network and unwrapped to UV coordinate space. 

\begin{figure}[t]
	\centerline{\includegraphics[width=0.5\textwidth]{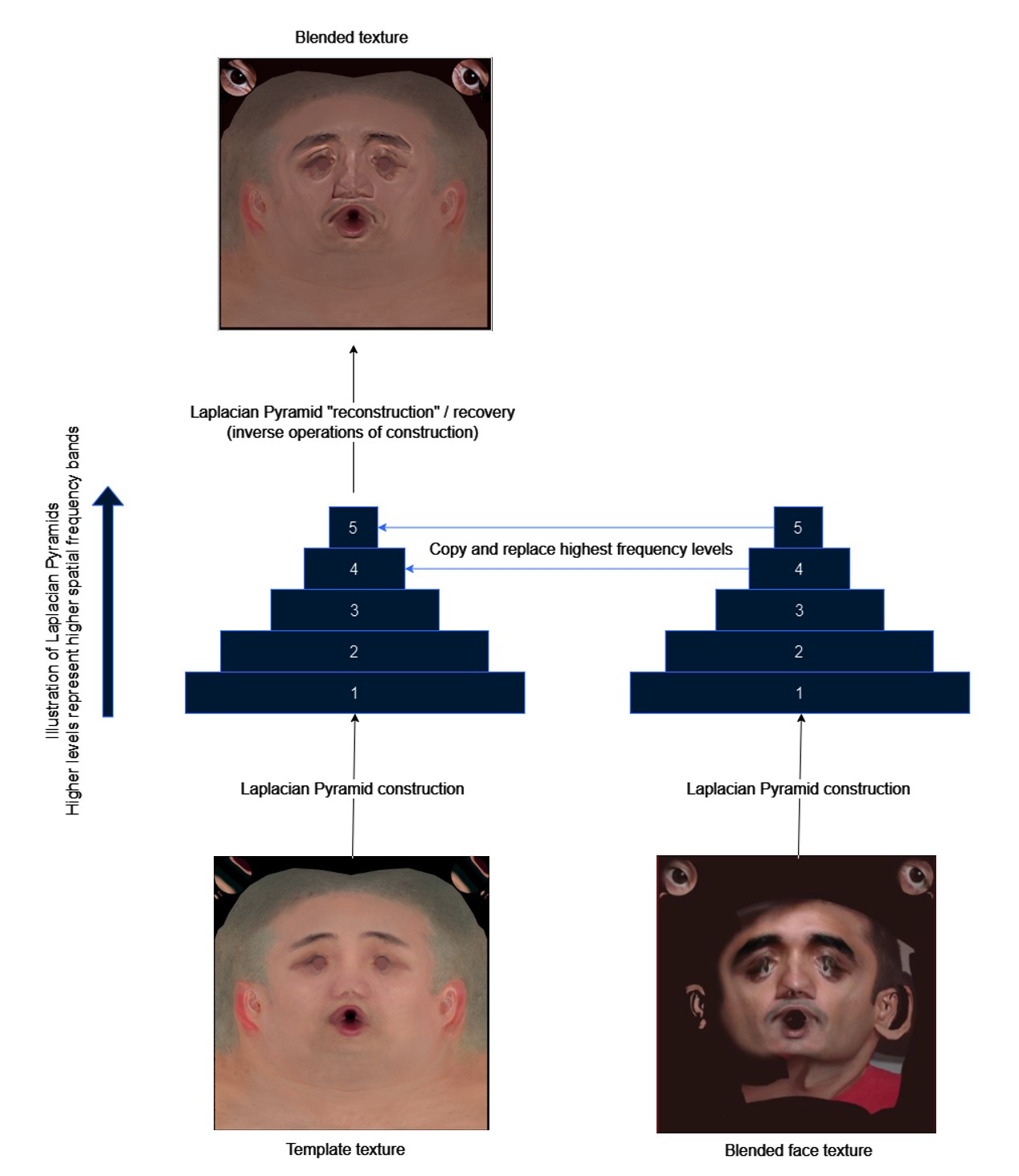}}    
	\caption{An illustration of our novel Laplacian pyramid blending procedure. It takes a facial texture generated from a set of unevenly or directionally illuminated source images, and a template facial texture with neutral illumination, and produces an evenly illuminated texture containing unique identifiable features from the source images.}	
	\label{fig:laplacian_blending}
\end{figure}

\noindent \textbf{Laplacian pyramid blending:} A naive texture blending fails when there is an uneven or directional illumination, resulting in incorrect shading. To solve this problem, we observe that that uniquely identifiable features of a face are encoded in high spatial frequencies whereas intensity gradients due to lighting conditions are encoded in low spatial frequencies. Therefore, a solution to the problem is to transfer only the high frequency content from a facial texture generated from a set of unevenly illuminated images of a user's face onto an evenly illuminated template facial texture. We implement this solution using Laplacian pyramids. As shown in Figure \ref{fig:laplacian_blending}, given an unevenly illuminated facial texture and an evenly illuminated template facial texture, we replace the high-frequency levels of the Laplacian Pyramid of the template texture with that of the given facial texture. The resulting pyramid is reconstructed to a new texture. This resulting texture contains the identifiable facial features of the original facial texture while having the even illumination of the template texture. Using the same technique, we also generate a normal displacement map for the avatar using a template normal map from \cite{hifi3dface}, allowing for modeling fine bumps and pores in the skin. 

\noindent \textbf{Texture refinement network:} The generation pipeline must work with source images captured on consumer-grade laptop webcams which may be low-resolution or noisy and capture limited skin details. Figure \ref{fig:texture_refinement_network} shows the proposed self-supervised training pipeline for a texture refinement model. To train this network, ground-truth images are obtained from the FFHQ dataset. To simulate the degradation caused by a low-quality laptop webcam, we add degradation consisting of: Gaussian blur, downsampling, Gaussian noise, nonlocal-means filtering, and bi-linear upsampling back to the original size. The degradation parameters are sampled randomly at train time within the pre-determined range specified in Fig. \ref{fig:texture_refinement_network}. This synthetically degraded image is the input to the network. The U-Net follows the same architecture as \cite{pix2pix} and is trained to produce a refined texture ("Refined UV map") using a combination of loss functions. A photometric loss computes the pixel-wise L1 distance. The adversarial loss uses a separate discriminator model to provide supervision. The LPIPS perceptual loss computes the similarity between the activations of two image patches for a pre-trained image network, which has been shown to match human perception well.

\noindent \textbf{Texture refinement with face restoration: }An alternative approach to synthesizing realistic details in the texture generated from low-quality webcam images is to pre-process the source images with pre-trained face restoration models such as GFPGAN. These models are trained on large-scale face image datasets on a wide range of synthetic degradation and can perform detail synthesis and super-resolution simultaneously. 

\section{Results and Evaluation}

\begin{figure*}[t]
	\centering
	\includegraphics[width=0.95\textwidth]{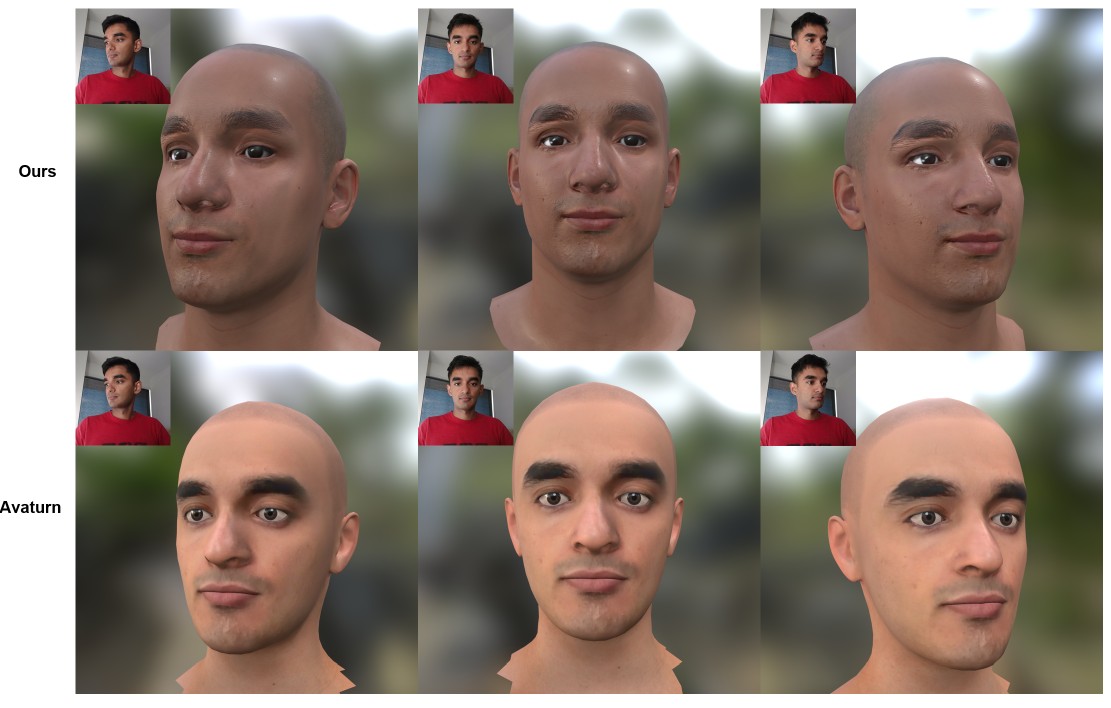}
	\caption{Comparison with Avaturn, a closed-source tool with similar computational cost.}	
	\label{fig:avaturn_comparison}
\end{figure*}

\noindent \textbf{Test cases:}
In order to test the effectiveness of our method at satisfying the requirements and tackling the challenges specified in section \ref{sec:intro}, we capture a set of test images as follows. Each set of images consists of a left, front and right view of the user, and all images are captured on a 1080p webcam on a commodity Lenovo ThinkPad laptop, with the following considerations:  (i) well-lit images with frontal illumination, (ii) well-lit images with directional illumination across the face, (iii) poorly-lit (backlit) images with directional illumination across the face, and images with the user sitting far away from the camera with directional illumination across the face.

\noindent \textbf{Qualitative Evaluations:}
Figure \ref{fig:results_main} shows high-fidelity renderings made in Blender of the avatars generated from the multi-view image test sets described above. The texture refinement based on the face restoration model has been applied in these examples, since we observed it to yield more realistic textures. As seen in test cases \textit{b)} and \textit{d)}, our Laplacian pyramid blending works well in removing soft directional illumination across the face. In test case \textit{c)}, while the illumination on the side of the face is successfully removed, the harsh highlight on the side of the nose results in a visible artifact. Test case \textit{c)} also shows how our texture refinement helps improve quality when the user's face occupies fewer pixels in the source image as a result of them sitting far from the camera.

\noindent \textbf{Quantitative metrics:} Considering that we are targeting user grade compute resources, subjective visual quality of generated and rendered avatars is an important part of our evaluation. We compare our method to \href{https://avaturn.me/}{Avaturn}, a recently developed closed-source proprietary avatar generation platform, as it is the closest to our method in terms of computational cost and target use case.
\begin{table*}[!t]
	\centering
	\caption{Quantitative comparison of our method with Avaturn. Lower is better.}
	\begin{tabular}{@{}c|ccc@{}}
		\toprule
		Method                    & w/o LP blending & Avaturn & with LP blending \textbf{(ours)}   \\ \midrule
		Brightness Symmetry Error & 0.0624        & 0.0179  & \textbf{0.0119} \\
		Face Identity Emb. Distance & -        & 18.21 & \textbf{15.68 } \\ \bottomrule
	\end{tabular}
	\label{table:bse}
\end{table*}

\begin{figure}[]
	\centering
	\includegraphics[width=0.45\textwidth]{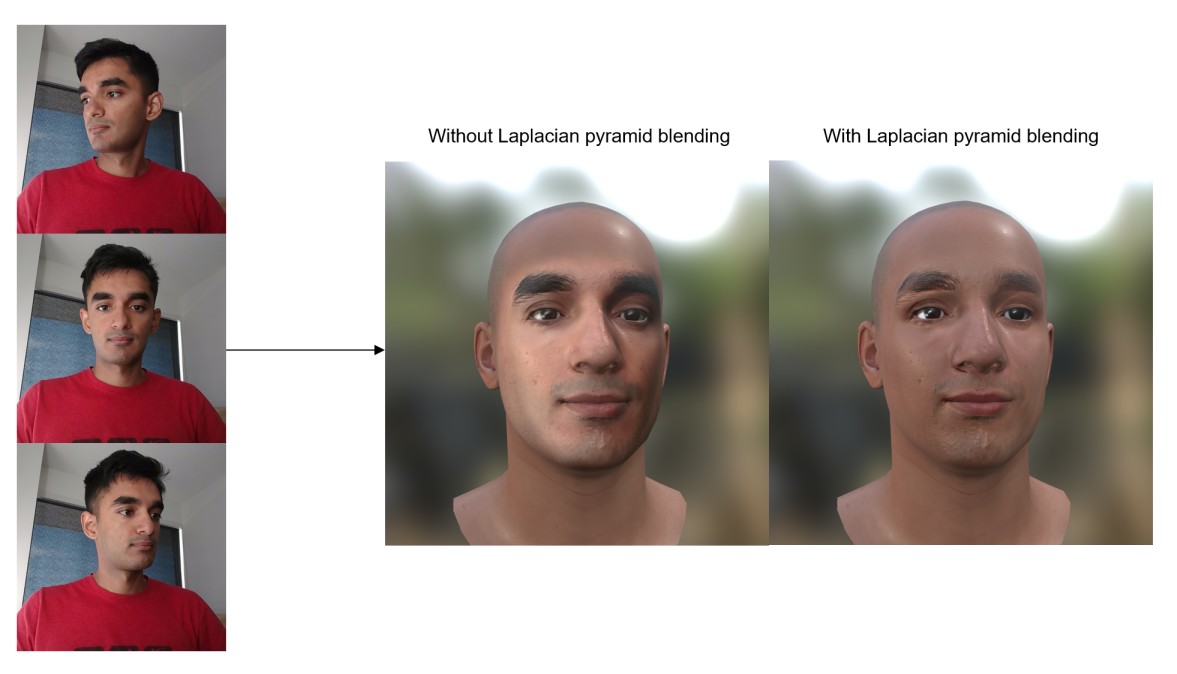}
	\caption{With our novel Laplacian pyramid blending procedure, although the source images contain directional illumination across the face, the resulting texture is neutrally illuminated, allowing the avatar to be accurately re-illuminated in downstream applications.}
	\label{fig:laplacian_comparison}
\end{figure}

\textbf{Brightness Symmetry Error: } Following \cite{ffhq-uv}, for a given texture UV map $T$, we define the BS Error as:
\[BSE(T) = \vert\vert \mathcal{B}_\alpha (T^Y) - \mathcal{F}_h (\mathcal{B}_\alpha (T^Y)) \vert\vert_1\]
where $T^Y$ denotes the $Y$ channel of $T$ in YUV color space; $\mathcal{B}(\cdot)$ denotes Gaussian blur with kernel size $\alpha$, set to 55 empirically; $\mathcal{F}(\cdot)$ denotes horizontal flipping. Table \ref{table:bse} shows that our Laplacian pyramid blending method achieves a significant reduction in BS Error on our unevenly illuminated test cases, and also achieves lower BS Error than Avaturn.

\textbf{Perceptual Distance: } In order to evaluate the quality improvement, we compare the LPIPS perceptual distance of the degraded and refined textures to their corresponding high-quality textures. Applying this distance metric to a subset of the FFHQ dataset that was not used in training the model, we get an average degraded-to-HQ perceptual distance of \textbf{0.092} and an average refined-to-HQ perceptual distance of \textbf{0.035}. This means that the refined textures are perceptually closer to the original high-quality textures than the degraded textures.

\section{Conclusions}
In this work, we propose and implement an avatar generation pipeline designed to work with source images from consumer laptop webcams, deal with directional illumination across a user's face, be computationally inexpensive to run, and results in an avatar that can be animated and rendered in real-time. Our method shows that classical 3D and image processing methods such as 3DMMs and Laplacian pyramids combined with modern methods based on GANs and differentiable rendering can produce visually appealing avatars from a few low-quality images in resource-constrained environments. The results show evidence of the novelty and quality of our work.


\bibliographystyle{IEEEbib}
\bibliography{references_og}

\end{document}